\documentclass[conference, 10pt]{IEEEtran}
\IEEEoverridecommandlockouts
\usepackage[a4paper, total={184mm,235mm}]{geometry}

\usepackage{cite}
\usepackage{amsmath,amssymb,amsfonts}
\usepackage{graphicx}
\usepackage{textcomp}
\usepackage{xcolor}
\def\BibTeX{{\rm B\kern-.05em{\sc i\kern-.025em b}\kern-.08em
    T\kern-.1667em\lower.7ex\hbox{E}\kern-.125emX}}
\usepackage{booktabs} 
\usepackage{multirow}
\usepackage{xcolor}
\usepackage[super]{nth}
\usepackage{tikz}
\newcommand*\circled[1]{\tikz[baseline=(char.base)]{
            \node[shape=circle,draw,inner sep=0.2pt] (char) {#1};}}
\newcommand*\circledB[1]{\tikz[baseline=(char.base)]{
            \node[shape=circle,fill,inner sep=0.2pt] (char) {\textcolor{white}{#1}};}}

\usepackage{soul}
\usepackage{enumitem}
\usepackage{algorithmic}
\usepackage{algorithm}
\usepackage{array}
\newcolumntype{?}{!{\vrule width 1.5pt}}
\newcommand{\add}[1]{\textcolor{black}{#1}}

\usepackage{setspace}

\usepackage{fancyhdr}
\fancypagestyle{firstpage}{
  \fancyhf{}
  \chead{\small To appear at the 2022 International Joint Conference on Neural Networks (IJCNN), \\ the 2022 IEEE World Congress on Computational Intelligence (WCCI), July 2022, Padova, Italy.}
  \cfoot{\thepage}
}

\begin{document}

\title{\huge lpSpikeCon: Enabling Low-Precision Spiking Neural Network Processing for Efficient Unsupervised Continual Learning on Autonomous Agents
\vspace{-0.3cm}}

\author{\IEEEauthorblockN{ Rachmad Vidya Wicaksana Putra\IEEEauthorrefmark{1}, Muhammad Shafique\IEEEauthorrefmark{2}}
\IEEEauthorblockA{\IEEEauthorrefmark{1}\textit{Institute of Computer Engineering}, 
\textit{Technische Universit\"at Wien (TU Wien)},
Vienna, Austria\\
\IEEEauthorrefmark{2}\textit{Division of Engineering}, 
\textit{New York University Abu Dhabi (NYUAD)},
Abu Dhabi, United Arab Emirates\\
Email: rachmad.putra@tuwien.ac.at, muhammad.shafique@nyu.edu}
\vspace{-0.8cm}
}

\maketitle
\pagestyle{plain}
\thispagestyle{firstpage}

\begin{spacing}{0.936}
\begin{abstract}
Recent advances have shown that Spiking Neural Network (SNN)-based systems can efficiently perform unsupervised continual learning due to their bio-plausible learning rule, e.g., Spike-Timing-Dependent Plasticity (STDP). 
Such learning capabilities are especially beneficial for use cases like autonomous agents (e.g., robots and UAVs) that need to continuously adapt to dynamically changing scenarios/environments, where new data gathered directly from the environment may have novel features that should be learned online. 
Current state-of-the-art works employ high-precision weights (i.e., 32 bit) for both training and inference phases, which pose high memory and energy costs thereby hindering efficient embedded implementations of such systems for battery-driven mobile autonomous systems. 
On the other hand, precision reduction may jeopardize the quality of unsupervised continual learning due to information loss. 
Towards this, we propose lpSpikeCon, a novel methodology to enable low-precision SNN processing for efficient unsupervised continual learning on resource-constrained autonomous agents/systems. 
Our lpSpikeCon methodology employs the following key steps: (1) analyzing the impacts of training the SNN model under unsupervised continual learning settings with reduced weight precision on the inference accuracy; (2) leveraging this study to identify SNN parameters that have a significant impact on the inference accuracy; and (3) developing an algorithm for searching the respective SNN parameter values that improve the quality of unsupervised continual learning. 
The experimental results show that our lpSpikeCon can reduce weight memory of the SNN model by 8x (i.e., by judiciously employing 4-bit weights) for performing online training with unsupervised continual learning and achieve no accuracy loss in the inference phase, as compared to the baseline model with 32-bit weights across different network sizes.
\end{abstract}
\end{spacing}

\vspace{-0.3cm}
\begin{IEEEkeywords}
Spiking neural networks, SNNs, unsupervised learning, continual learning, memory efficiency, energy efficiency, autonomous agents, embedded systems.
\end{IEEEkeywords}

\begin{spacing}{0.936}
\section{Introduction}
\label{Sec_Intro}

\begin{figure}[t]
\centering
\includegraphics[width=\linewidth]{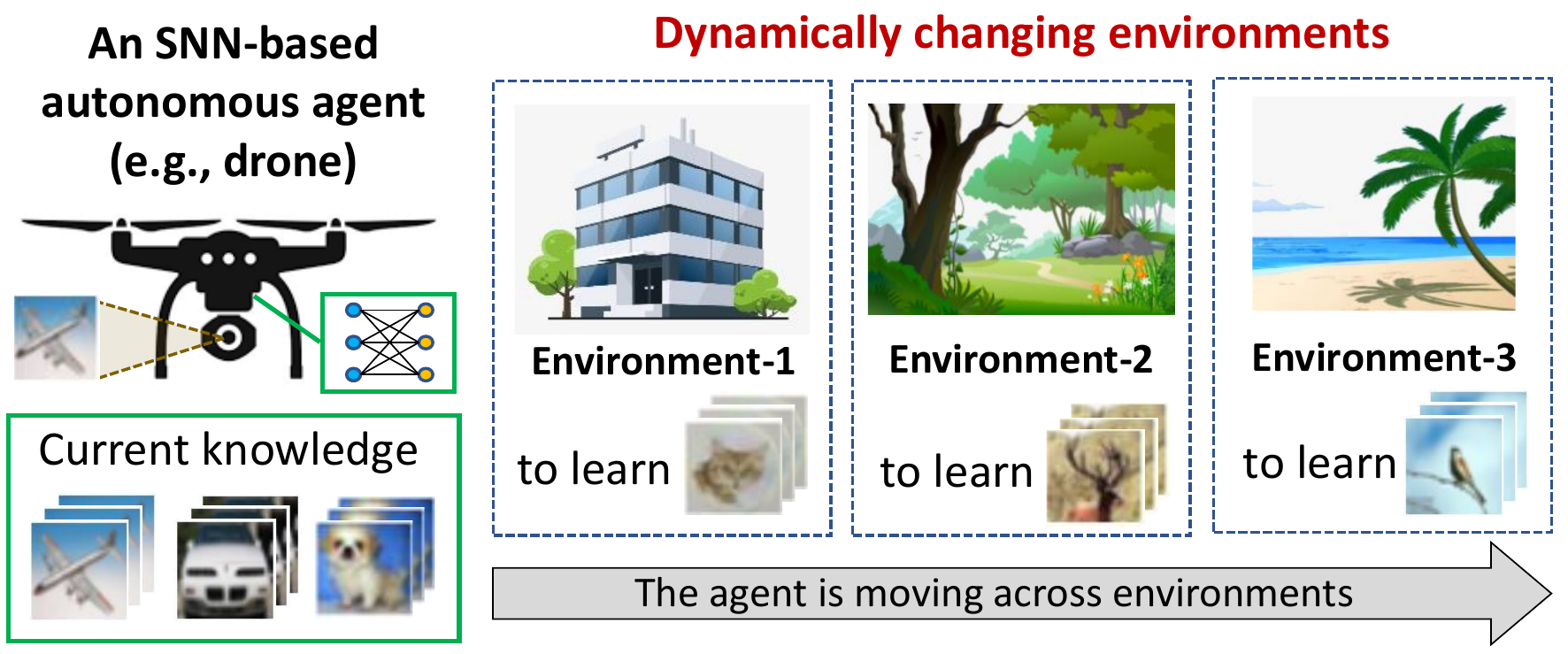}
\vspace{-0.6cm}
\caption{The SNN-based autonomous agent needs to perform training online using unsupervised continual learning strategies to update the knowledge, thereby adapting to dynamically changing environments in an efficient manner.}
\label{Fig_DynamicEnv}
\vspace{-0.4cm}
\end{figure}

With the advances of neuromorphic computing, SNN-based systems have shown potential for having efficient learning capabilities due to their bio-plausible learning rules, like the Spike-Timing-Dependent Plasticity (STDP)~\cite{Ref_Pfeiffer_DLSNN_FNINS18, Ref_Tavanaei_DLSNN_Neunet18, Ref_Davies_Loihi_MM18, Ref_Hazan_SOMSNN_IJCNN18, Ref_Saunders_STDPpatch_IJCNN18, Ref_Hazan_LMSNN_AMAI19, Ref_Saunders_LCSNN_NeuNet19, Ref_Putra_FSpiNN_TCAD20}.
However, the knowledge learned from the offline training process may become obsolete over time and in unpredictable operational conditions, which leads to low accuracy at run time. 
This is crucial when SNN systems are deployed in the dynamically changing scenarios/environments, i.e., where new data collected directly from the operational environments have novel information/features that should be learned online~\cite{Ref_Panda_ASP_JETCAS18}\cite{Ref_Putra_SpikeDyn_DAC21}; see Fig.~\ref{Fig_DynamicEnv}. 
Moreover, this data is typically unlabeled and may not be proportionally distributed, thereby making it difficult for the SNN systems to learn different tasks/classes~\cite{Ref_Allred_ForcedFiring_IJCNN16}. 
To address these issues, recent works have proposed strategies for enabling \textit{the unsupervised continual learning in SNNs}~\cite{Ref_Allred_ForcedFiring_IJCNN16, Ref_Panda_ASP_JETCAS18, Ref_Allred_CFN_FNINS20, Ref_Putra_SpikeDyn_DAC21}, which are beneficial for use cases like autonomous agents (e.g., robots and UAVs) that need to continuously adapt to dynamically changing environments~\cite{Ref_Putra_SpikeDyn_DAC21}. 
However, autonomous agents typically have tight constraints (e.g., available memory), thereby requiring an efficient embedded implementation of SNN systems. 

\textbf{Targeted Problem}: 
\textit{If and how can we efficiently implement SNN systems with unsupervised continual learning capabilities on autonomous agents with tight memory constraints, while maintaining the inference accuracy close to that of the baseline implementation.}
An efficient solution to this problem will enable highly memory- and power/energy-efficient autonomous agents that are adaptive to unpredictable dynamic environments.

\subsection{The State-of-the-Art and Their Limitations}
\label{Sec_Intro_SoA}

The research for unsupervised continual learning in SNN systems is still at an early stage. 
Therefore, the state-of-the-art works still focus on improving the quality of learning, while considering relatively simple datasets (e.g., MNIST)~\cite{Ref_Allred_ForcedFiring_IJCNN16, Ref_Panda_ASP_JETCAS18, Ref_Allred_CFN_FNINS20, Ref_Putra_SpikeDyn_DAC21}. 
These works employ high-precision weights (i.e., 32 bits) for both training and inference phases to achieve high accuracy, but incur high memory and energy costs, thereby hindering the implementation of such SNN systems for battery-powered autonomous agents.
Towards this, quantization is a potential technique for efficiently reducing the memory footprint of SNNs, and hence the energy consumption~\cite{Ref_Putra_FSpiNN_TCAD20}\cite{Ref_Putra_QSpiNN_IJCNN21}.
However, \textit{the impacts of weight quantization on the accuracy of unsupervised continual learning in SNN systems have not been explored yet}.
To show the limitations of state-of-the-art and the targeted problem, we perform an experimental case study in Section~\ref{Sec_Intro_Challenges}.

\begin{figure}[t]
\centering
\includegraphics[width=0.95\linewidth]{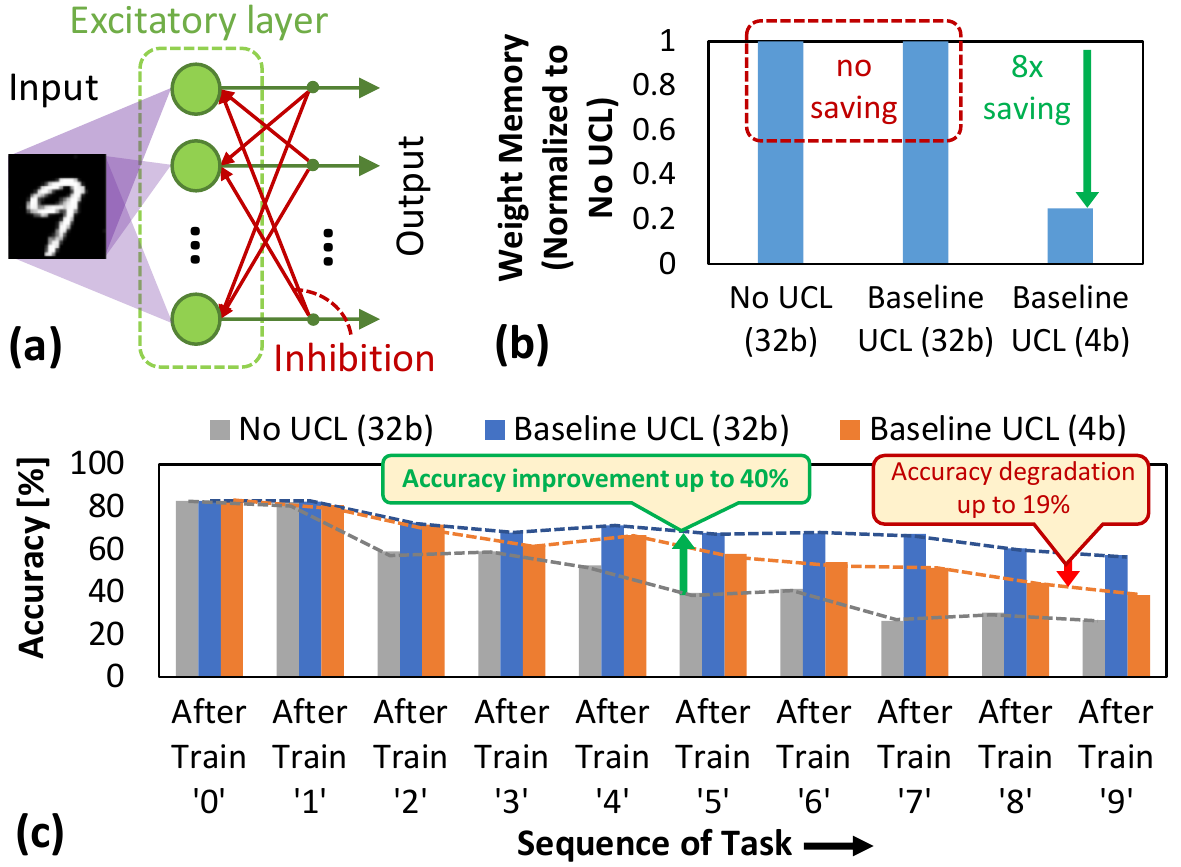}
\vspace{-0.3cm}
\caption{(a) The SNN architecture that supports unsupervised continual learning, i.e., a single-layer fully-connected network. (b) Weight memory of a 400-neuron network with different learning conditions: No Unsupervised Continual Learning (No UCL) with 32-bit weights; Baseline UCL with 32-bit weights, adapted from~\cite{Ref_Putra_SpikeDyn_DAC21}; and Baseline UCL with 4-bit weights, using quantization. (c) Accuracy of a 400-neuron network with different learning conditions.
\add{In this work, we consider the UCL algorithm from the work of~\cite{Ref_Putra_SpikeDyn_DAC21}}.}
\label{Fig_CaseStudy}
\vspace{-0.6cm}
\end{figure}

\subsection{Case Study and Scientific Challenges}
\label{Sec_Intro_Challenges}

For the case study, we perform experiments that provide dynamic scenarios to the network by feeding consecutive tasks/ classes using training samples, train the network accordingly, then evaluate the trained network using the test samples for tasks/classes that have been fed so far.  
Following are the steps of experiments using the MNIST dataset.
\begin{itemize}[leftmargin=*]
    \item First, we feed a stream of training samples for digit-0, and train the network accordingly. 
    Then, we evaluate the trained network using the test samples for digit-0.
    \item Second, we repeat the above steps but for training digit-1, and testing digit-0 and digit-1.
    \item The above steps are repeated but for training another digit, and testing the digits that have been learned so far, until all 10 digits in MNIST are used for training and testing.
\end{itemize}

Our experiments consider the network architecture shown in Fig.~\ref{Fig_CaseStudy}(a) and different learning conditions: (1) No Unsupervised Continual Learning (No UCL); (2) Baseline UCL with 32-bit weights, adapted from~\cite{Ref_Putra_SpikeDyn_DAC21}; and (3) Baseline UCL with 4-bit weights.  
Further details of the experimental setup are presented in Section~\ref{Sec_EvalMethod}.
The experimental results are shown in Fig.~\ref{Fig_CaseStudy}(b)-(c), from which we make the following key observations.
\begin{itemize}[leftmargin=*]
    \item The unsupervised continual learning improves the accuracy under dynamic scenarios, due to its carefully crafted weight potentiation/depression strategy to learn new features, while retaining old yet important ones.
    \item Reduction of weight precision can significantly save the SNN weight memory, e.g., reducing precision from 32-bit to 4-bit weights enables 8x weight memory saving. 
    However, it may degrade the quality of unsupervised continual learning due to knowledge/information loss.
    \item A network with 4-bit weights and continual learning may achieve higher accuracy than a network with 32-bit weights but no continual learning, thereby showing the potential of memory reduction for a network under dynamic scenarios.
\end{itemize}

These observations highlight the following key challenges to devise an efficient solution for the targeted problem.
\begin{itemize}[leftmargin=*]
    \item \textit{Quantization should be performed judiciously to remove non-significant information in each weight}, hence retaining most of the important information and maintaining the learning quality (i.e., high accuracy).
    \item \textit{The solution should employ simple yet effective enhancements} to compensate for the information loss due to weight quantization, thereby enabling energy-efficient learning. 
\end{itemize}

\subsection{Our Novel Contributions}
\label{Sec_Intro_Novelty}

\begin{figure}[h]
\vspace{-0.3cm}
\centering
\includegraphics[width=\linewidth]{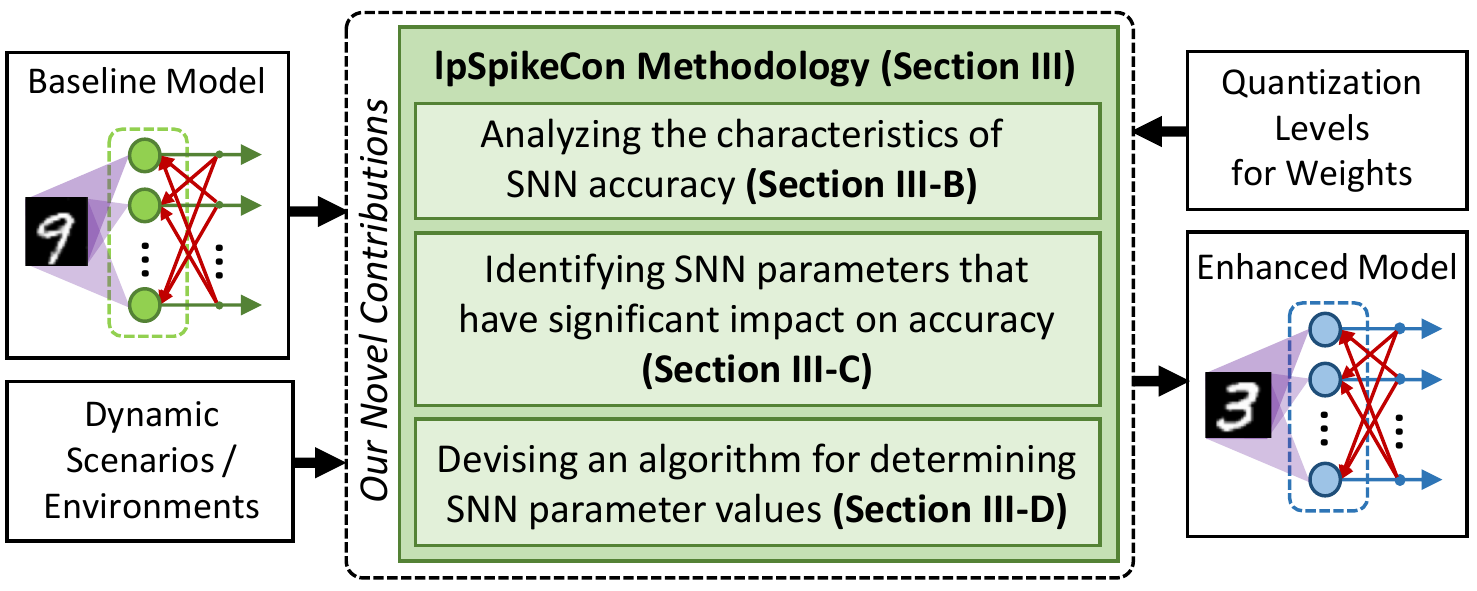}
\vspace{-0.7cm}
\caption{The lpSpikeCon methodology, highlighting the novel components.}
\label{Fig_NovelContrib}
\vspace{-0.3cm}
\end{figure}

To address the above challenges, we propose \textbf{lpSpikeCon}, a novel methodology that enables \underline{l}ow-\underline{p}recision \underline{Spik}ing neural network processing for \underline{e}fficient unsupervised \underline{Con}tinual learning on autonomous agents/systems. 
To the best of our knowledge, this work is the first effort that aims at \textit{reducing the weight precision of SNNs, while maintaining the quality of unsupervised continual learning} under dynamic scenarios. 
Following are the key steps of our lpSpikeCon methodology (see an overview in Fig.~\ref{Fig_NovelContrib}).
\begin{itemize}[leftmargin=*]
    \item \textbf{Analyzing the characteristics of SNN accuracy profiles of each given task under different quantization levels}, when the given SNN employs unsupervised continual learning.  
    \item \textbf{Identifying the SNN parameters that have significant impact on the accuracy}. 
    It leverages the accuracy analysis to determine SNN parameters and their adjustment rules to get better neuronal dynamics for unsupervised continual learning, and hence accuracy. 
    \item \textbf{Devising an algorithm for determining parameter values} that effectively improve the learning quality. 
    It leverages the parameter adjustment rules for guiding the algorithm to refine the parameter values for achieving high accuracy in dynamic scenarios/environments. 
\end{itemize}

\textbf{Key Results:} 
We evaluate our lpSpikeCon methodology using a Python-based framework~\cite{Ref_Hazan_BindsNET_FNINF18} on a multi-GPU machine.
The experimental results show that our lpSpikeCon improves the quality of unsupervised continual learning for a quantized SNN through effective parameter adjustments, thereby achieving no accuracy loss in the inference while reducing the weight memory (e.g., by 8x when the SNN employs 4-bit weights), as compared to the non-quantized SNN (i.e., with 32-bit weights). 

\section{Background}
\label{Sec_Background}

\subsection{Spiking Neural Networks (SNNs)}
\label{Sec_Background_SNNs}

SNNs have the potential for energy-efficient processing in both the training and inference phases due to their highly bio-plausible computation model, i.e., spike-based information and computation~\cite{Ref_Maass_SNN_NeuNet97, Ref_Putra_ReSpawn_ICCAD21, Ref_Shafique_EdgeAI_ICCAD21, Ref_Putra_SparkXD_DAC21}. 
Each SNN model is formed by several design components, i.e., network architecture, neuron and synapse models, spike coding, as well as learning rule~\cite{Ref_Putra_FSpiNN_TCAD20}.

\begin{figure}[h]
\vspace{-0.2cm}
\centering
\includegraphics[width=0.92\linewidth]{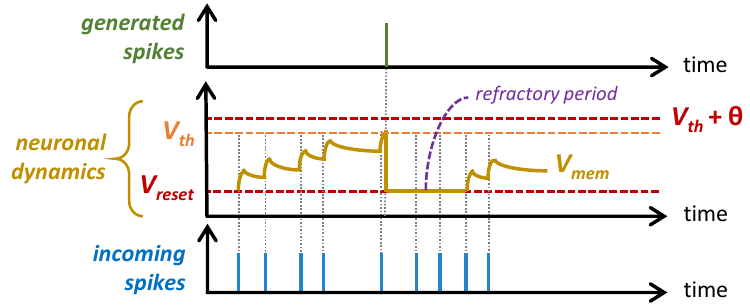}
\vspace{-0.3cm}
\caption{Overview of the neuronal dynamics of a LIF model.}
\label{Fig_LIFmodel}
\vspace{-0.3cm}
\end{figure}

In this work, we employ the network architecture/topology of Fig.~\ref{Fig_CaseStudy}(a), as it has been widely used for SNNs with unsupervised continual learning capabilities~\cite{Ref_Allred_ForcedFiring_IJCNN16, Ref_Panda_ASP_JETCAS18, Ref_Putra_SpikeDyn_DAC21}. 
This network is trained so that each excitatory neuron can recognize a specific task/class.
For spike/information coding, several coding schemes have been proposed in the literature, such as temporal, rank-order, rate, phase, and burst coding \cite{Ref_Gautrais_SpikeCoding_Bio98, Ref_Kayser_PhaseCoding_Neuron09, Ref_Thorpe_RankOrder_Springer98,Ref_Park_BurstSNN_DAC19}. 
Here, we employ the rate coding to convert information into spike trains, since it can achieve high accuracy in unsupervised learning settings~\cite{Ref_Putra_SoftSNN_arXiv22}. 
For the neuron model, we employ the Leaky Integrate-and-Fire (LIF) model, as it incurs the lowest computational complexity while providing highly bio-plausible neuronal dynamics, as compared to other neuron models~\cite{Ref_Izhikevich_CompareModels_TNN04}. 
The neuronal dynamics of the LIF model are shown in Fig.~\ref{Fig_LIFmodel}, and explained in the following. 
\begin{itemize}[leftmargin=*]
    \item LIF neuron increases the membrane potential ($V_{mem}$) \add{each time there is} an incoming spike, and otherwise, the $V_{mem}$ decreases exponentially with the rate of $V_{decay}$.
    \item If the $V_{mem}$ reaches the membrane threshold potential ($V_{th}$), the LIF neuron generates a spike, then the $V_{mem}$ goes back to the reset potential ($V_{reset}$), and the neuron state enters the refractory period ($T_{ref}$). 
    \item To evenly distribute the spiking activity across all excitatory neurons (i.e., homeostasis), the membrane threshold potential is defined as $V_{th} = V_{th}+\theta$ (so-called \textit{adaptive membrane threshold potential})~\cite{Ref_Diehl_STDPmnist_FNCOM15} with $\theta$ denotes the adaptation potential. The $\theta$ is added to $V_{th}$ each time the neuron generates a spike, and otherwise, $V_{th}$ decreases with the rate of $\theta_{decay}$. 
\end{itemize}
Meanwhile, the synapse is modeled by the conductance and weight ($w$). 
The conductance is increased by $w$ when there is learning activity (i.e., weight potentiation). 
Depending upon the learning rule, the $w$ may be decreased with the rate of weight decay ($w_{decay}$) if there is no learning activity (i.e., weight depression).
For the learning rule, we employ the \add{pair-based STDP learning}, which is adapted from~\cite{Ref_Putra_SpikeDyn_DAC21}. 

\subsection{Quantization}
\label{Sec_Background_Quantization}

\begin{figure}[h]
\vspace{-0.3cm}
\centering
\includegraphics[width=\linewidth]{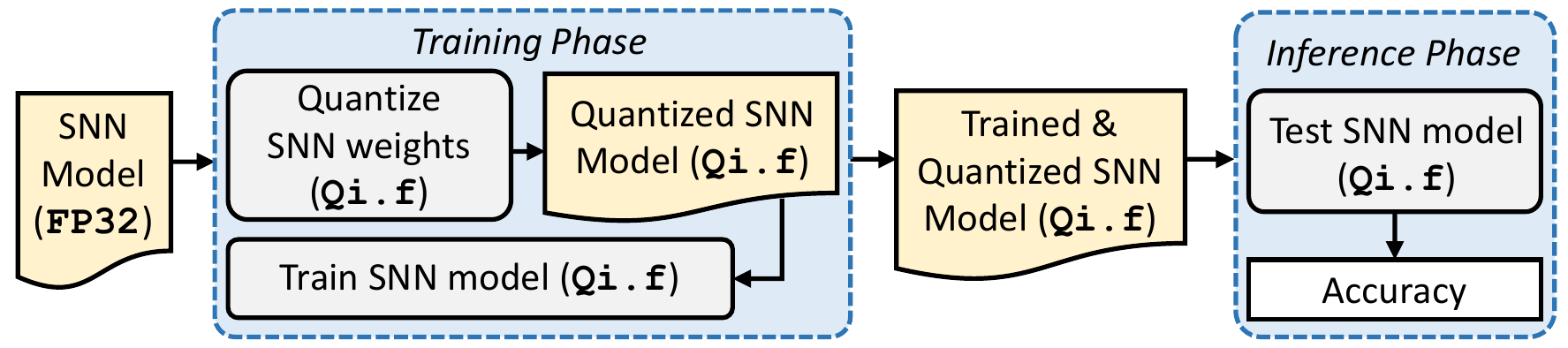}
\vspace{-0.6cm}
\caption{Overview of the quantization scheme.}
\label{Fig_QuantScheme}
\vspace{-0.3cm}
\end{figure}

Quantization is a prominent technique to reduce the size of an SNN model (i.e., weight memory) that incurs a relatively low overhead, since it only needs to reduce the data precision~\cite{Ref_Putra_QSpiNN_IJCNN21}.
It can be represented in a fixed-point format (\texttt{Qi.f}), which consists of 1 sign bit, \texttt{i} integer bits, \texttt{f} fractional bits, and follows the 2's complement format.
Therefore, the \texttt{Qi.f} format has a range of representable values of $[-2^{\texttt{i}}, 2^{\texttt{i}}-2^{\texttt{-f}}]$ with a precision of $\epsilon = 2^{\texttt{-f}}$. 
%
In the quantization process, a rounding scheme is required, and we consider the simple and widely used one, i.e., \textit{truncation}~\cite{Ref_Hopkins_Rounding_RSTA20}\cite{Ref_Gupta_DLPrecision_ICML15}. 
Truncation simply keeps \texttt{f} bits and removes the remaining bits from the fractional part. 
The output of fixed-point quantization for a given real number $x$ with configuration \texttt{Qi.f} is defined as $T(x,\texttt{Qi.f}) = \lfloor x \rfloor$. 
To simulate the unsupervised continual learning under quantized weights, we perform quantization during the training phase using the simulated quantization approach~\cite{Ref_Krishnamoorthi_Whitepaper_arXiv18}; see Fig.~\ref{Fig_QuantScheme}. 

\section{The lpSpikeCon Methodology}
\label{Sec_lpSpikeCon}

\subsection{Overview}
\label{Sec_lpSpikeCon_Overview}

\begin{figure*}[t]
\centering
\includegraphics[width=0.92\linewidth]{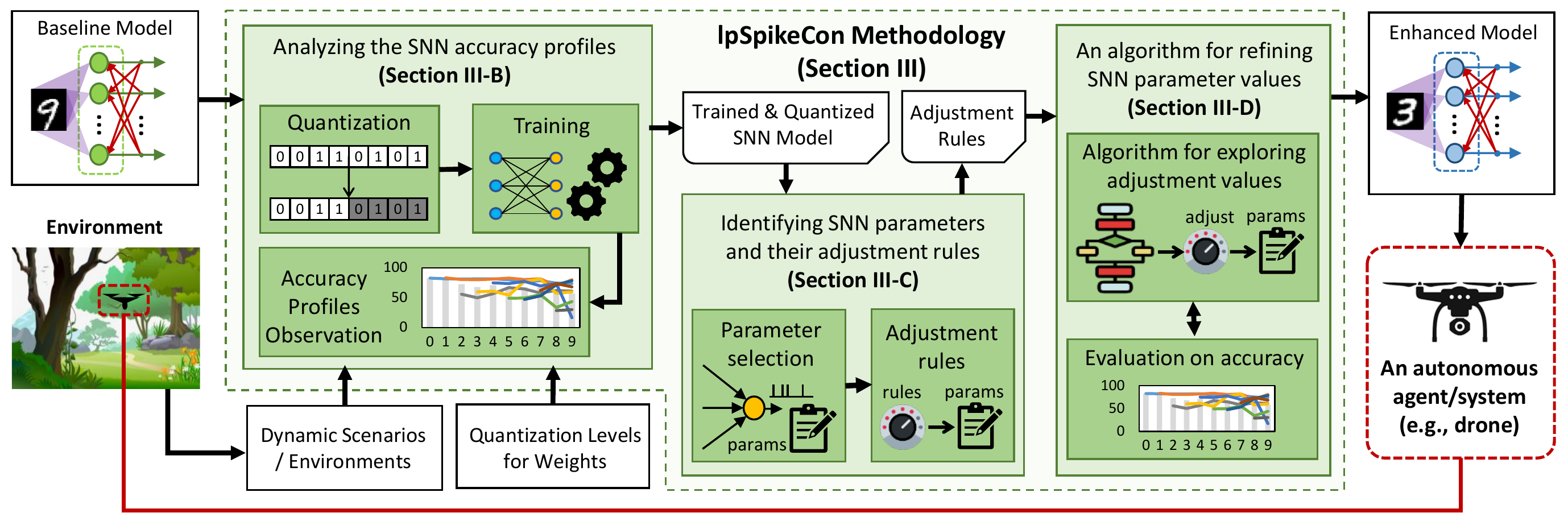}
\vspace{-0.3cm}
\caption{The overview of our lpSpikeCon methodology, whose novel steps are shown in green boxes.}
\label{Fig_lpSpikeCon}
\vspace{-0.5cm}
\end{figure*}

Our lpSpikeCon methodology employs the following key steps (see an overview in Fig.~\ref{Fig_lpSpikeCon}). 
\begin{itemize}[leftmargin=*]
    \item \textbf{Analyzing the SNN accuracy profiles (Section~\ref{Sec_lpSpikeCon_Analysis})} by performing the following mechanisms. 
    \begin{itemize}
        \item Quantizing the weights of a given network. 
        \item Training the network with quantized weights using the unsupervised continual learning under dynamic scenarios.
        \item Observing the accuracy profiles of each given task/class under different quantization levels. 
    \end{itemize}
    \item \textbf{Identifying the SNN parameters and their adjustment rules for unsupervised continual learning (Section~\ref{Sec_lpSpikeCon_Identify})} by performing the following mechanisms.
     \begin{itemize}
        \item Leveraging the accuracy analysis to select SNN parameters that have a significant impact on accuracy. 
        \item Devising the adjustment rules for the selected parameters to derive better neural dynamics (e.g., membrane threshold potential) for unsupervised continual learning.
    \end{itemize}
    \item \textbf{Devising an algorithm for refining SNN parameter values for unsupervised continual learning (Section~\ref{Sec_lpSpikeCon_Algo})}. 
    It is performed by leveraging the adjustment rules to gradually increase/decrease parameter values that improve the learning quality under quantized weights.
\end{itemize}

\add{The lpSpikeCon-enhanced SNN systems are then scheduled to update their offline-trained knowledge regularly at run time, e.g., based on the user-defined duration of an inference mode. 
The update is performed through online training using data gathered from the operational environments.
After completing the training mode, the systems are back to the inference mode. 
In this manner, the SNN systems are expected to adapt better to diverse operational environments than the offline-trained ones.} 

\subsection{Analyzing the SNN Accuracy Profiles}
\label{Sec_lpSpikeCon_Analysis}

To devise a lightweight solution that enables efficient unsupervised continual learning for a quantized SNN, \textit{it is important to understand the characteristics of SNN accuracy for each given task/class under different quantization levels and dynamic scenarios}.
To do this, we perform the following steps.
\begin{itemize}[leftmargin=*]
    \item We quantize the weights of a given SNN model using the truncation approach, as described in Section~\ref{Sec_Background_Quantization}.
    \item Then, we train the quantized model using unsupervised continual learning under dynamic scenarios. To do this, we feed consecutive tasks/classes using training samples, train the model accordingly, then evaluate the trained model using the test samples for tasks/classes that have been fed so far; see the overview in Alg.~\ref{Alg_DynamicScenarios}.
    \item Afterwards, we observe the profiles of inference accuracy for each given task/class under different levels of weight quantization, to understand the sources of accuracy degradation and get insights on how to address it.
\end{itemize}

\begin{algorithm}[hbtp]
\footnotesize
\caption{Training and testing under dynamic scenarios}
\label{Alg_DynamicScenarios}
\begin{algorithmic}[1]
\renewcommand{\algorithmicrequire}{\textbf{INPUT:}}
\renewcommand{\algorithmicensure}{\textbf{OUTPUT:}}
\REQUIRE \textbf{(1)} SNN model ($model_{in}$); 
\textbf{(2)} Dataset ($D$): dataset for class-$i$ ($D[i]$), training set for class-$i$ ($D[i].train\_set$), testing set for class-$i$ ($D[i].test\_set$); \\
\ENSURE \textbf{(1)} Trained SNN model ($model_{out}$); \textbf{(2)} Accuracy ($acc$)\\
\vspace{0.1cm}
\renewcommand{\algorithmicrequire}{\textbf{BEGIN}}
\renewcommand{\algorithmicensure}{\textbf{END}}
\REQUIRE \hspace{0.1cm} \\   
    \textbf{Initialization}: \\
      \STATE $model_{out} = model_{in}$; \\
    \textbf{Process}: \\
      \STATE $task = class(D)$
      \FOR{$i \in task$} 
        \STATE $model_{out} \leftarrow$ train$(model_{out},  D[i].train\_set)$; \\
        \FOR{$k = 0$ to $i$}
          \STATE $acc[i,k] =$ test$(model_{out},  D[k].test\_set)$; \\
        \ENDFOR
      \ENDFOR 
    \RETURN $model_{out}$; \\
\ENSURE 
\end{algorithmic} 
\end{algorithm}
\setlength{\textfloatsep}{2pt}

\begin{figure*}[t]
\centering
\includegraphics[width=0.95\linewidth]{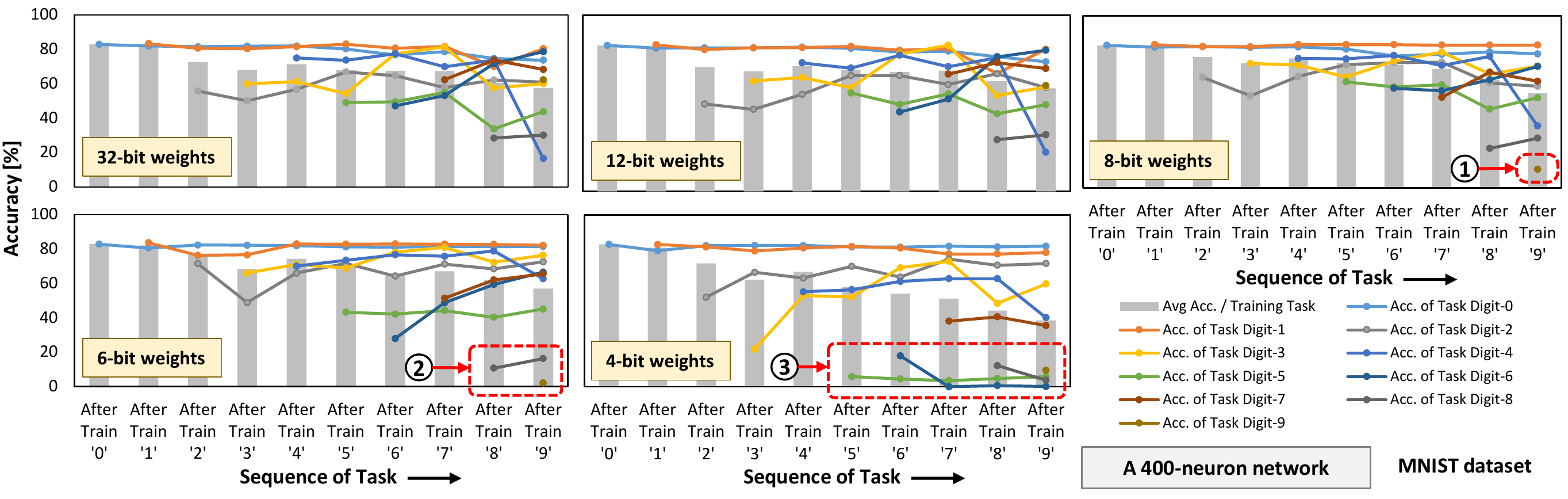}
\vspace{-0.3cm}
\caption{The accuracy profiles of a 400-neuron network on MNIST dataset under different levels of weight precision (i.e., 32, 12, 8, 6, and 4 bits) and dynamic scenarios. 
The colored line represents the accuracy for each task/class throughout the consecutive training phases from digit-0 to digit-9.
The grey bar represents the average accuracy after each training phase of a task/class.}
\label{Fig_AnalysisAccuracy}
\vspace{-0.5cm}
\end{figure*}

The experimental results are presented in Fig.~\ref{Fig_AnalysisAccuracy}, from which we make the following key observations.
\begin{itemize}[leftmargin=*]
    \item In general, reduced weight precision degrades the accuracy of unsupervised continual learning due to information loss. 
    The accuracy degradation is noticeable especially for tasks that are learned at the later training sequence, as shown by labels \circled{1}, \circled{2}, and \circled{3} in Fig.~\ref{Fig_AnalysisAccuracy}.
    \item Lower weight precision also leads to lower accuracy for more tasks. 
    For instance, a model with 4-bit weights suffers from very low accuracy ($\leq 20\%$) on four tasks, i.e., digit-6, digit-7, digit-8, and digit-9 (see label \circled{3} in Fig.~\ref{Fig_AnalysisAccuracy}), while a model with 8-bit weights suffers from very low accuracy only on task digit-9 (see label \circled{1} in Fig.~\ref{Fig_AnalysisAccuracy}).
\end{itemize}

\smallskip
These observations lead to several insights for devising an efficient solution for the targeted problem, as follows. 
\begin{itemize}[leftmargin=*]
    \item Reduced weight precision has a less memory capacity for storing unique information from new tasks, thereby making the quantized SNN model difficult to recognize novel features from new tasks. 
    \item Different SNN models with different levels of weight precision have different accuracy profiles. Therefore, such models require specialized settings for achieving high accuracy. 
\end{itemize}
The above discussion indicates that, \textit{the potential solution is by employing parameter adjustments that provide more available memories for storing novel information from new tasks, while considering specialized settings for different levels of weight precision}. 
In this manner, costly additional components and/or operations can be avoided, which leads to efficient learning. 
 
\subsection{Identifying SNN Parameters and Their Adjustment Rules}
\label{Sec_lpSpikeCon_Identify}

Analysis in Section~\ref{Sec_lpSpikeCon_Analysis} suggests that the efficient solution is by employing parameter adjustments for the given quantized SNN model. 
Therefore, \textit{it is important to identify SNN parameters that have a significant impact on accuracy, and their effective adjustment rules for improving learning quality of the quantized model}.

\smallskip
\textbf{SNN Parameters:} 
To identify SNN parameters that should be adjusted for better learning quality, we leverage the analysis from Section~\ref{Sec_lpSpikeCon_Analysis}.
The analysis shows that the reduced weight precision makes the model difficult to learn tasks that appear at the later training sequence. 
This means that the weight bits are strongly associated with the previously learned tasks, and not flexible enough to change their context to another task. 
To address this, \textit{we adjust the neuronal dynamics so that the synaptic plasticity becomes more flexible for learning new tasks, especially when low weight precision is employed}. 
To do this, we adjust two parameters to change the flexibility of neuronal dynamics for learning activity: \textit{adaptive membrane threshold potential} ($V_{th}$) and \textit{weight decay rate} ($w_{decay}$). 
This selection is based on the following reasons.
\begin{itemize}[leftmargin=*]
    \item $V_{th}$ determines how far $V_{mem}$ should be increased to generate a spike, which then triggers weight potentiation for learning. 
    The distance between $V_{th}$ and $V_{mem}$ is inversely proportional to the flexibility for learning activity.
    \item $w_{decay}$ determines how fast each synaptic weight is depressed for removing the learned information, and providing available memory for storing novel information from new tasks.  
\end{itemize}

\begin{figure}[t]
\centering
\includegraphics[width=0.95\linewidth]{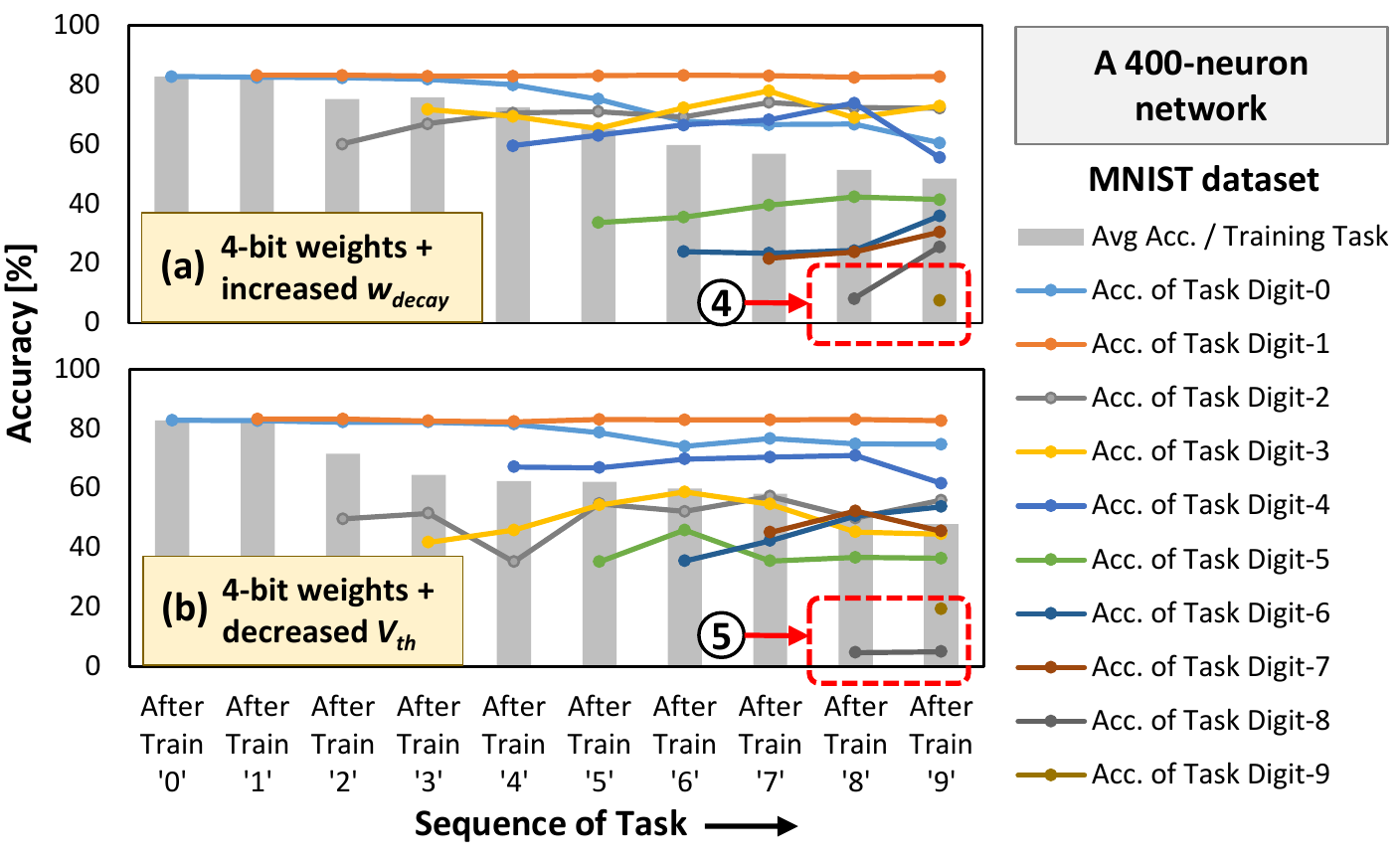}
\vspace{-0.3cm}
\caption{The accuracy profiles of a 400-neuron network with 4-bit weights on the MNIST dataset under dynamic scenarios and different parameter adjustments. (a) ``Increased $w_{decay}$" by using \text{baseline} $w_{decay}+0.09$. (b) ``Decreased $V_{th}$" by using $V_{th}+$ (\text{baseline} $\theta-0.2$).}
\label{Fig_AnalysisAdjust}
\end{figure}

\smallskip
\textbf{Adjustment Rules:}
To properly adjust the selected SNN parameters (i.e., $V_{th}$ and $w_{decay}$) for achieving better learning quality of a quantized SNN model, we propose the following \textit{parameter adjustment rules}. 
\begin{itemize}[leftmargin=*]
    \item In the non-quantized model, $V_{th}$ is set with a value that prevents catastrophic forgetting\footnote{Catastrophic forgetting means that the system learns information from new data, but quickly forgets the previously learned ones~\cite{Ref_Chen_Lifelong_MLP18,Ref_Parisi_CLL_NeuNet19,Ref_McCloskey_CI_Elsevier89}}. 
    Therefore, \textit{to make $V_{th}$ more suitable for the quantized model, its value should be less or equal ($\leq$) than $V_{th}$ of the non-quantized model}.
    In this manner, each neuron is expected to reach $V_{th}$ faster and to generate spikes more frequently, which triggers learning activity for any incoming tasks.  
    \item In the non-quantized model, $w_{decay}$ is set with a value that provides available memory for storing learned information from new tasks. 
    Therefore, \textit{to make $w_{decay}$ more suitable for the quantized model, its value should be greater or equal ($\geq$) than $w_{decay}$ of the non-quantized model}. 
    In this manner, each weight is expected to decay faster, hence providing available memory for storing learned features from any incoming tasks. 
\end{itemize}
To justify these rules, we perform an experimental case study to see the impact of our adjustment rules, i.e., ``decreased $V_{th}$" and ``increased $w_{decay}$". 
Experimental results are provided in Fig.~\ref{Fig_AnalysisAdjust}.
These results show that our adjustment rules improve the quality of unsupervised continual learning for both ``decreased $V_{th}$" and ``increased $w_{decay}$" cases, since each case has less number of tasks with very low accuracy, i.e., $\leq 20\%$ (see labels \circled{4} and \circled{5} in Fig.~\ref{Fig_AnalysisAdjust}), as compared to the model with 4-bit weights and baseline continual learning (see label \circled{3} in Fig.~\ref{Fig_AnalysisAccuracy}).

\subsection{An Algorithm for Refining SNN Parameter Values}
\label{Sec_lpSpikeCon_Algo}

To properly adjust the values of the selected SNN parameters, a systematic mechanism is required. 
Toward this, \textit{we propose an algorithm that leverages our adjustment rules to refine the selected SNN parameter values ($V_{th}$ and $w_{decay}$) for achieving better learning quality, especially under dynamic scenarios.}
This algorithm is developed based on the following ideas, and the detailed steps are provided in Alg.~\ref{Alg_AdjustmentSteps}.
\begin{itemize}[leftmargin=*]
    \item The range of values for each selected SNN parameter ($V_{th}$ or $w_{decay}$) should be defined for guiding the design space exploration. 
    The lower-bound value for $V_{th}$ is $V^l_{th}$, while the upper-bound value for $w_{decay}$ is $w^u_{decay}$.
    \item The accuracy for each task ($acc_{task}$) should not be less or equal ($\leq$) than the defined value. 
    Therefore, if we consider $acc_{low}$ as the borderline of low accuracy, then the acceptable accuracy for each task is defined as follows.
    \begin{equation}
    \small
    acc_{task} > acc_{low}
    \label{Eq_LowAccuracy}
    \end{equation}
    \item The average accuracy of the lpSpikeCon-enhanced SNN model (${acc^*_{avg}}$) should be within an acceptable accuracy loss. 
    Therefore, if we consider ${acc_{avg}}$ as the average accuracy of the non-quantized SNN model with baseline unsupervised continual learning, and $acc_{loss}$ as the acceptable accuracy loss, then the ${acc^*_{avg}}$ is defined as follows.
    \begin{equation}
    \small
    acc^*_{avg} \geq (acc_{avg} - acc_{loss})
    \label{Eq_AvgAccuracy}
    \end{equation}
    \end{itemize}
    
\begin{algorithm}[hbtp]
\footnotesize
\caption{Adjustment steps for SNN parameter values}
\label{Alg_AdjustmentSteps}
\begin{algorithmic}[1]
\renewcommand{\algorithmicrequire}{\textbf{INPUT:}}
\renewcommand{\algorithmicensure}{\textbf{OUTPUT:}}
\REQUIRE \textbf{(1)} SNN: baseline non-quantized model ($model_{in}$), weight decay rate ($w_{decay}$), upper-bound of $w_{decay}$ ($w^u_{decay}$); 
\textbf{(2)} Exploration variables: investigated/evaluated model ($model_{eval}$), increasing step for $w_{decay}$ ($step\_w$), decreasing step for $V_{th}$ ($step\_V_{th}$); 
\textbf{(3)} Functions: accuracy for each task ($acc_{task}$), average accuracy of the given model ($acc$), acceptable accuracy loss ($acc_{loss}$);
\ENSURE Trained SNN model ($model_{out}$); \\
\vspace{0.1cm}
\renewcommand{\algorithmicrequire}{\textbf{BEGIN}}
\renewcommand{\algorithmicensure}{\textbf{END}}
\REQUIRE \hspace{0.1cm} \\   
    \textbf{Initialization}: \\
      \STATE $model_{out} = model_{in}$; \\
    \textbf{Process}: \\
      \FOR{$(p = w_{decay}; \;\; p \leq w^u_{decay}; \;\; p = p+step\_w)$} 
        \FOR{$(q = V_{th}; \;\; q \geq V^l_{th}; \;\; q = q-step\_V_{th})$} 
          \STATE $model_{eval} = model_{in}$; \\
          \STATE update the values of selected SNN parameters;
          \STATE perform \textit{training} and \textit{test} on $model_{eval}$ using Alg.~\ref{Alg_DynamicScenarios} \\
          \IF{$(acc_{task}(model_{eval}) > acc_{low}) \;\; \text{\textbf{and}} \;\; (acc(model_{eval}) \geq (acc(model_{in}) - acc_{loss}))$}  
            \IF{($acc(model_{eval}) \geq acc(model_{out})$)}
              \STATE $model_{out} = model_{eval}$ 
            \ENDIF
          \ENDIF
        \ENDFOR
      \ENDFOR 
    \RETURN $model_{out}$; \\
\ENSURE 
\end{algorithmic} 
\end{algorithm}
\setlength{\textfloatsep}{6pt}

\section{Evaluation Methodology}
\label{Sec_EvalMethod}

\begin{figure}[h]
\vspace{-0.2cm}
\centering
\includegraphics[width=\linewidth]{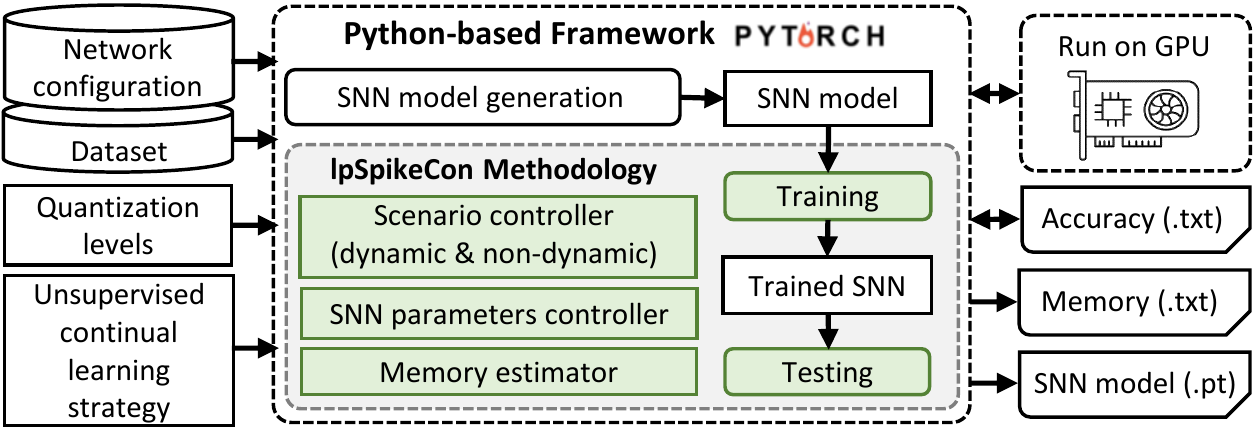}
\vspace{-0.7cm}
\caption{The experimental setup for evaluating our lpSpikeCon methodology.}
\label{Fig_EvalMethod}
\vspace{-0.3cm}
\end{figure}

Fig.~\ref{Fig_EvalMethod} shows the experimental setup for evaluating lpSpikeCon methodology. 
We employ a Python-based framework~\cite{Ref_Hazan_BindsNET_FNINF18} \add{that runs on} a multi-GPU machine (i.e., Nvidia RTX 2080 Ti) and an Embedded-GPU machine (i.e., Nvidia Jetson Nano) to perform evaluations on different platforms with different memory and compute capabilities.
We employ a single-layer fully-connected network shown in Fig.~\ref{Fig_CaseStudy}(a) with different network sizes (i.e., 200 and 400 neurons), \add{since it has shown the capabilities for performing unsupervised continual learning under resource- and power-constrained embedded platforms~\cite{Ref_Putra_SpikeDyn_DAC21}}. 
We consider MNIST dataset as workload, since it has been widely used \add{for evaluating unsupervised continual learning in the SNN} community~\cite{Ref_Allred_ForcedFiring_IJCNN16, Ref_Panda_ASP_JETCAS18, Ref_Allred_CFN_FNINS20, Ref_Putra_SpikeDyn_DAC21}. 
For the baseline unsupervised continual learning, we consider the learning strategy from~\cite{Ref_Putra_SpikeDyn_DAC21}.
The evaluation is performed under both the non-dynamic and dynamic scenarios. 
\textit{Non-dynamic scenarios} are provided by feeding the network with training samples whose tasks/classes are randomly distributed. 
It aims at simulating conventional offline training where all
training samples are already available. 
Meanwhile, \textit{dynamic scenarios} are provided by feeding the network with consecutive tasks/classes, where each task/class has the same number of samples, and without re-feeding previous tasks/classes. 
It aims at simulating an extreme condition where the deployed system receives training tasks in a consecutive manner from the environment.

\section{Results and Discussion}
\label{Sec_Results}

\subsection{Maintaining Accuracy under Quantized Weights}
\label{Sec_Results_Accuracy}

\begin{figure*}[t]
\centering
\includegraphics[width=\linewidth]{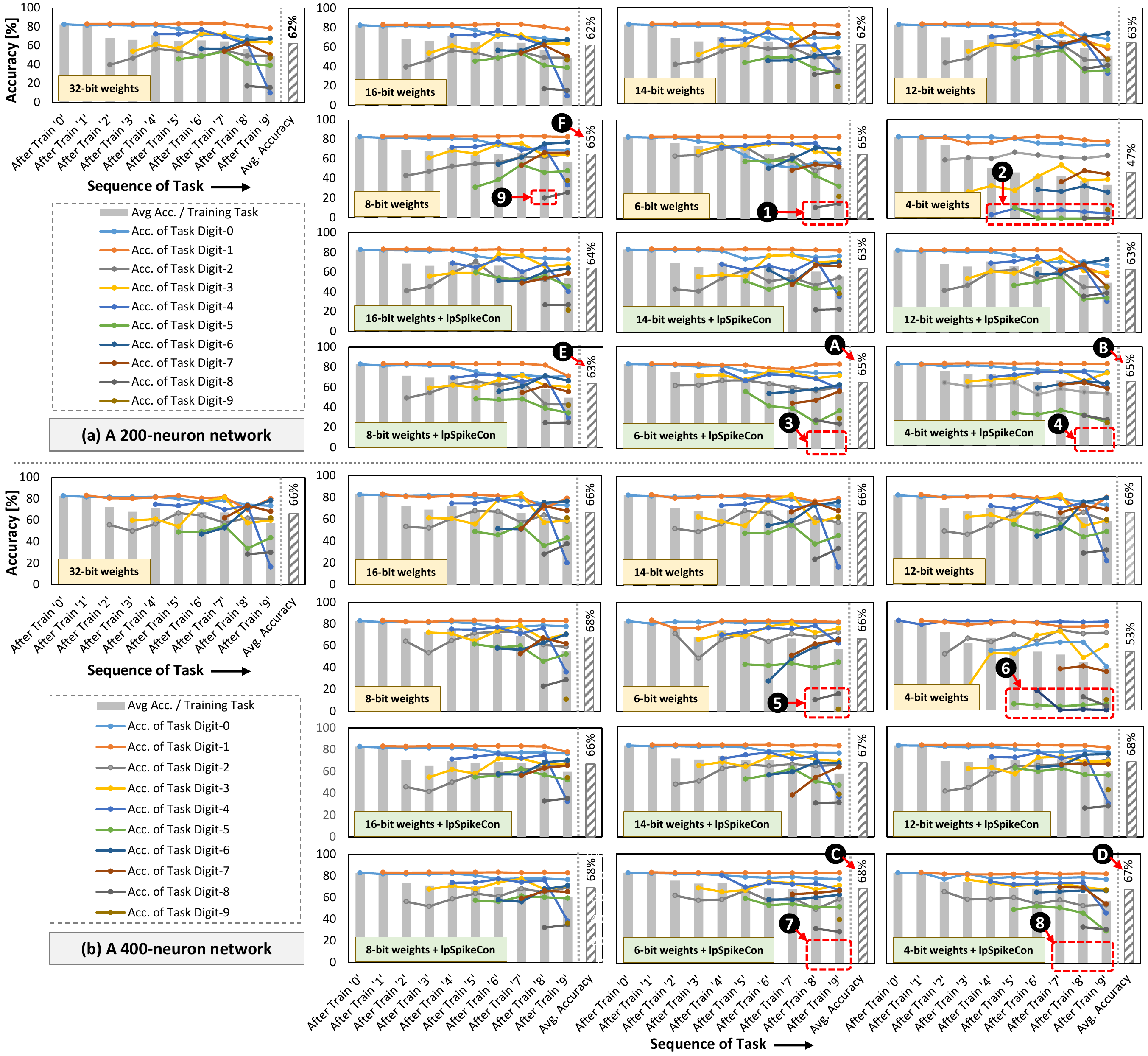}
\vspace{-0.7cm}
\caption{The accuracy profiles of (a) a 200-neuron network, and (b) a 400-neuron network, under different levels of weight precision and dynamic scenarios. The colored line represents the inference accuracy for each task/class throughout the consecutive training phases. The grey-colored bar represents the average accuracy after each training phase of a task/class. The pattern-coded bar represents the average accuracy for all evaluated tasks/classes.}
\label{Fig_Results_AccuracyDynY}
\vspace{-0.5cm}
\end{figure*}

\textbf{Dynamic Scenarios:}
We evaluate accuracy considering different network sizes (i.e., 200 and 400 neurons) and different weight precision levels (i.e., 32, 16, 14, 12, 8, 6, and 4 bits), and the experimental results are provided in Fig.~\ref{Fig_Results_AccuracyDynY}.
These results show that lower weight precision leads to lower accuracy for more recognition tasks due to information loss. 
For instance, in a 200-neuron network, 6-bit weights lead to very low accuracy (i.e., $\leq 20\%$) on one task (see label \circledB{1}), while 4-bit weights lead to very low accuracy on four tasks (see label \circledB{2}).
Such patterns are also observed in a 400-neuron network, as shown by labels \circledB{5} and \circledB{6}.

The experimental results also show that our lpSpikeCon methodology can improve the accuracy of the quantized SNNs, which can be observed in two aspects.
First, the lpSpikeCon-enhanced SNNs do not suffer from very low accuracy (i.e., $\leq 20\%$) for recognizing any tasks/classes; see labels \circledB{3} and \circledB{4} for a 200-neuron network, and labels \circledB{7} and \circledB{8} for a 400-neuron network. 
Second, the lpSpikeCon-enhanced SNNs also achieve no accuracy loss on average when compared to the non-quantized SNNs with baseline unsupervised continual learning, across different network sizes. 
For instance, in a 200-neuron network, average accuracy for the 6-bit and 4-bit weights with lpSpikeCon is 65\%, which is slightly higher than the 32-bit weights with baseline learning (i.e., 62\%); see labels \circledB{A} and \circledB{B}. 
\add{A similar pattern} is also observed in a 400-neuron network, as the average accuracy for the 6-bit and 4-bit weights with lpSpikeCon are 68\% and 67\% respectively, which are slightly higher than the 32-bit weights with baseline learning (i.e., 66\%); see labels \circledB{C} and \circledB{D}. 
These accuracy improvements are due to proper adjustments on the selected SNN parameters (i.e., $V_{th}$ and $w_{decay}$). 
These adjustments trigger the neuronal dynamics in the SNN system to quickly provide available memory for learning new tasks through the ``increased $w_{decay}$" approach, and trigger learning activity for any incoming tasks (including the new tasks) through the ``decreased $V_{th}$" approach.
From the results, we also observe that in a certain case, an lpSpikeCon-enhanced model may have lower average accuracy than the non-enhanced model under the same weight precision.
For instance, in the 200-neuron network with 8-bit weights, our lpSpikeCon-enhanced model achieves 63\% accuracy (see label~\circledB{E}), while the non-enhanced one achieves 65\% (see label~\circledB{F}).
The reason is that, the lpSpikeCon considers the borderline of low accuracy ($acc_{low}$) as a constraint to determine the parameter adjustments and the output model. 
Therefore, since the non-enhanced model has lower accuracy than the defined $acc_{low}$ for one task, this model is not considered as the solution (see label~\circledB{9}). 

\begin{figure}[h]
\centering
\includegraphics[width=\linewidth]{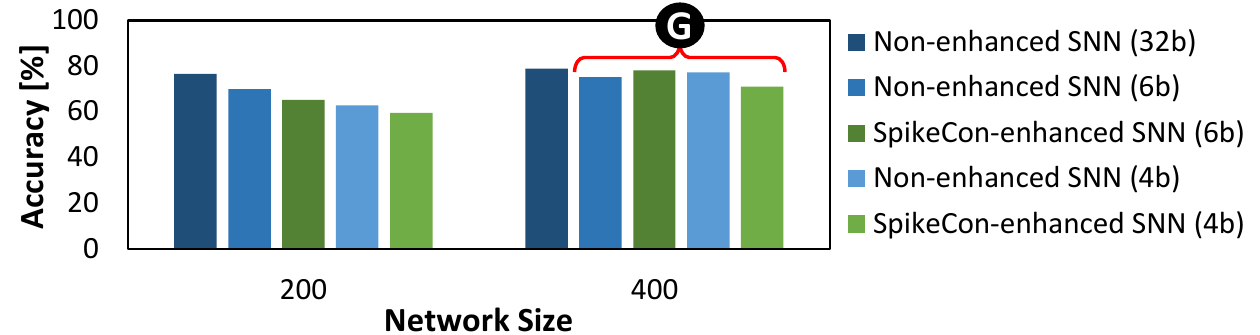}
\vspace{-0.7cm}
\caption{The accuracy of (a) a 200-neuron network, and (b) a 400-neuron network, under different levels of weight precision and non-dynamic scenarios. 
Here, the non-enhanced SNN refers to the model that employs baseline unsupervised continual learning.}
\label{Fig_Results_AccuracyDynN}
\vspace{-0.1cm}
\end{figure}

\textbf{Non-Dynamic Scenarios:}
We evaluate accuracy considering different network sizes (i.e., 200 and 400 neurons) and different weight precision levels (i.e., 32, 16, 14, 12, 8, 6, and 4 bits), and the experimental results are provided in Fig.~\ref{Fig_Results_AccuracyDynN}.
These results show that, our lpSpikeCon-enhanced SNNs can achieve comparable accuracy to the non-enhanced SNN counterparts (i.e., SNNs which employ baseline unsupervised continual learning under the same weight precision).
For instance, in the 400-neuron network with 6-bit weights, our lpSpikeCon-enhanced SNN achieves 78\% accuracy and the non-enhanced one achieves 75\% accuracy; while in the 4-bit weights case, the lpSpikeCon-enhanced SNN achieves 71\% accuracy and the non-enhanced one achieves 77\% accuracy, as highlighted by label \circledB{G}. 
The reason is that, our lpSpikeCon methodology performs exploration for parameter adjustments within a range of values that is close to the baseline settings. 
In this manner, proper adjustment values can be found fast, and are expected to preserve good characteristics from the baseline settings of unsupervised continual learning, such as achieving high accuracy under non-dynamic scenarios.

\subsection{Weight Memory Savings for Efficient SNN Systems}
\label{Sec_Results_Memory}

\begin{figure}[h]
\vspace{-0.4cm}
\centering
\includegraphics[width=\linewidth]{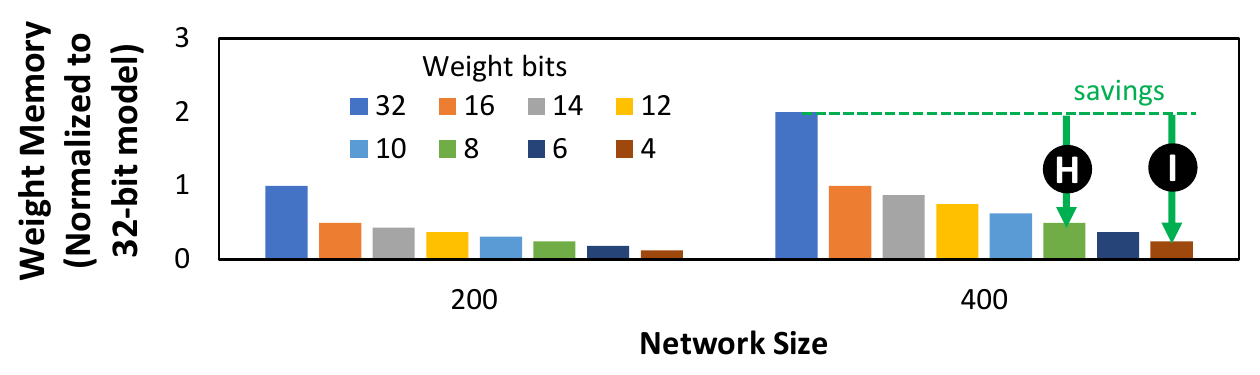}
\vspace{-0.7cm}
\caption{Weight memory requirements of SNN models with different sizes of network (i.e., 200 and 400 excitatory neurons) and different weight precision, which are normalized to the non-quantized SNN model (i.e., 32-bit weights).}
\label{Fig_Results_Memory}
\vspace{-0.1cm}
\end{figure}

Fig.~\ref{Fig_Results_Memory} shows the memory requirements of different SNN models under different levels of weight precision. 
These results show that weight quantization in our lpSpikeCon methodology can significantly decrease the memory footprint, since fewer bits are required to represent all weight parameters of an SNN model. 
For instance, a model with 8-bit weights reduces the weight memory by 4x (as shown by label~\circledB{H}), while a model with 4-bit weights reduces the weight memory by 8x (as shown by label~\circledB{I}), as compared to the non-quantized model which employs 32-bit weights. 
Reduced memory requirement is not only leading to a smaller area in hardware implementation, but \add{also} decreasing the number of (off-chip and on-chip) memory accesses. 
For instance, a non-quantized model requires a single DRAM-based off-chip memory access for obtaining a 32-bit weight, while a quantized model with 8-bit weight can access four weights from a single 32-bit DRAM access~\cite{Ref_Putra_DRMap_DAC20}\cite{Ref_Putra_ROMANet_TVLSI21}.
Memory access reduction is important to enable energy-efficient SNN-based autonomous agents mainly \add{for} two reasons.
First, memory accesses typically dominate the energy consumption of SNN systems, i.e., about 50\%-75\% of total energy consumption of an SNN accelerator~\cite{Ref_Krithivasan_SpikeBundle_ISLPED19}. 
Second, the online training with unsupervised continual learning requires frequent memory accesses for updating the weight values at run time.
Therefore, memory access reduction can significantly save the overall system energy, which is especially beneficial for memory- and energy-constrained autonomous agents/systems.

\add{The above results and discussion highlight our lpSpikeCon methodology provides several benefits as compared to the state-of-the-art works, including better learning quality under dynamic environments, smaller memory footprint, and higher energy efficiency. 
Furthermore, our lpSpikeCon can be extended further by incorporating network-specific parameters that have significant impacts on the accuracy.
For instance, networks with multiple layers may have additional parameters that should be considered for better adjustments towards adapting to new/unseen features.} 

\section{Conclusion}
\label{Sec_Conclusion}

We propose the lpSpikeCon methodology to enable low-precision SNN processing for efficient unsupervised continual learning on autonomous agents, through three key steps: (1) analysis of the SNN accuracy profiles, (2) identification of SNN parameters and their adjustment rules, and (3) refinements of parameter values for learning activity. 
As results, our lpSpikeCon significantly reduces weight memory of an SNN model, while maintaining accuracy in both dynamic and non-dynamic scenarios, as compared to the non-quantized model.
Therefore, our lpSpikeCon methodology may enable memory- and energy-efficient autonomous agents/systems that are adaptive to diverse operational environments.


\section*{Acknowledgment}
This work was partially supported by the NYUAD Center for Artificial Intelligence and Robotics (CAIR), funded by Tamkeen under the NYUAD Research Institute Award CG010. This work was also partially supported by the project ``eDLAuto: An Automated Framework for Energy-Efficient Embedded Deep Learning in Autonomous Systems”, funded by the NYUAD Research Enhancement Fund (REF).

\bibliographystyle{IEEEtran}
\bibliography{bibliography.bib}

\end{spacing}

\end{document}